\documentclass[letterpaper, 10 pt, conference]{ieeeconf}  

\IEEEoverridecommandlockouts                              

\usepackage{amssymb}  
\usepackage{graphicx}
\usepackage{amsmath}

\usepackage{algorithm}
\usepackage{algpseudocode}
\usepackage{adjustbox}
\usepackage{pdflscape}
\usepackage{geometry}
\usepackage{longtable}
\usepackage{array}
\usepackage{booktabs}
\usepackage{lscape}

\title{\LARGE \bf
Hybrid LLM-based Intelligent Framework for Robot Task Scheduling}

\author{
\authorblockN{Swayamjit Saha\textsuperscript{1,*}, Subhabrata Das\textsuperscript{2}, Haonan Duan\textsuperscript{3}, and Xiao-Yang Liu\textsuperscript{4}}
\authorblockA{\small
\textsuperscript{1}\textit{Department of Computer Science and Engineering}, \textit{Mississippi State University}, Starkville, MS, USA\\
\textsuperscript{2}\textit{Graduate School of Arts and Sciences, Columbia University}, New York, NY, USA\\
\textsuperscript{3}\textit{Consumer and Community Banking, JPMorgan Chase}, New York, NY, USA\\
\textsuperscript{4}\textit{Department of Data Science}, \textit{Columbia University}, New York, NY, USA\\
\textsuperscript{5}\textit{Department of Electrical Engineering}, \textit{Columbia University}, New York, NY, USA\\
\textsuperscript{1}ss4706@msstate.edu, \textsuperscript{2}sd2957@columbia.edu, \textsuperscript{3}hd2545@columbia.edu, \textsuperscript{4}xl2427@columbia.edu}\\
\textsuperscript{*}Corresponding author: Swayamjit Saha (ss4706@msstate.edu)
}

\begin{document}

\maketitle
\thispagestyle{empty}
\pagestyle{empty}

\begin{abstract}

This study introduces intelligent frameworks that use Large Language Models (LLMs) to improve task scheduling for construction robots. The LLM is fed with key data about the desired task, such as agent action abilities, and the desired end goal to be achieved. A well-balanced allocation strategy is developed, optimizing both time efficiency and resource utilization. Our system utilizes a Natural Language Processing interface to streamline communication with construction professionals and adapt in real-time to unexpected site conditions. We concurrently use two LLM agents, specifically generator (GPT-4) and  supervisor (Gemma 3/Llama 4/Mistral 7b) LLM agents to provide a more precise task schedule. We evaluate the proposed methodology using a straightforward scenario and provide metric scores to prove the efficacy of the frameworks. Our results highlight that the implementation of LLMs is crucial in construction operational tasks including robots.

\end{abstract}


\section{INTRODUCTION}

Construction is a vital and labor-intensive industry with a complicated working environment. Construction sites face a lot of variability, including the variable skill levels of the construction workforce, demanding robust, user-friendly and reliable technological solutions. In the paradigm of Construction 4.0 era, the applicablity of robots and autonomous systems has been sedentary. In order to aid the accelerating pace of construction engineering, it is critical to devise robotic systems to aid the construction workforce. Robotic systems increase efficacy in construction environments by aiding construction workers in its situational tasks \cite{c1}. However, flexible interaction between construction workers and robots is crucial, which could be achieved with natural language processing (NLP) and Large Language Models (LLMs). Our study explores the potential of LLM agents in robotic task allocation domain improving efficiency and accuracy of the robotic systems. 

The traditional method of schedule optimization and task allocation in construction has centered on creating specialized tools and software aimed at developing the most efficient schedules based on predefined task descriptions. Although, these softwares and tools offer valuable resources and platform for planning and scheduling construction tasks, they often incur the lack of the flexibility to adapt to the unprecedented changes that the construction environment faces. On the contrary, our research necessarily does not replace the aforementioned softwares and tools, but tends to investigate the new domain of task planning and schedule optimization using user-friendly natural language processing capabilities of LLM agents to facilitate human construction workers in a construction environment.

Our research presents a hybrid framework for the constructional task allocation and schedule optimization problem by using LLM agents, such as Gemma 3 by Google DeepMind \cite{c2}, Llama 4 by Meta AI \cite{c3}, Mistral 7b by Mistral AI SAS \cite{c4} and GPT-4 by OpenAI \cite{c5}. This paper presents a framework for task allocation and schedule optimization in construction settings by the following leveraging LLM agents. The main goal of this research is to improve the efficiency and flexibility of multi-agent systems in construction tasks. By taking into account complex factors like each agent's capabilities, estimated sub-task completion times and battery life optimization, the LLM agent creates an optimized strategy designed to reduce project duration while maximizing resource allocation and utilization. 


LLM agents have the ability to build logical relationships between varied tasks to achieve the desired goal \cite{c6,c15,c16}. However, the core of the mathematical base is built on a regression model, which is why LLMs sometimes lack in reasoning and predicting results in the future based on a set of logical constraints. Recent research studies have proved the implementation of multiple LLM agents in order to achieve more precise results on this domain, with the second LLM agent necessarily correcting or supervising the result from the first LLM agent \cite{c7}.

In practical scenario, interaction with the LLM agent to produce precise result involves quality of prompting. Recently, OpenAI \cite{c5} suggests dividing the prompt into two sets of information: background information and API. The API information involves the set of actions and commands for the LLM agent to process. Multimodal LLM agents have been proven to be useful for artificial intelligence (AI) robots in tasks involving but not limited to task allocation \cite{c8}, segmentation process \cite{c9} and so on. We discuss LLM agents used for robot task allocation and consecutively in the construction industry in the next section.


\section{BACKGROUND}

\subsection{LLM robot task allocation}

The latest advances in LLMs have already seen their way into the robot task allocation field. Wake, N. et al. \cite{c10} used a combination of GPT-4-based task planning and vision systems to extract spatial, temporal, and affordance information, enabling robots to execute tasks based on human demonstrations, while also noting the importance of human supervision due to occasional hallucinations in GPT-4.

Chalvatzaki, G. et al. \cite{c11} explored using GPT-2, a smaller large language model, for long-horizon robotic task planning by decomposing tasks into subgoals grounded in a scene graph for executable plans. The results show that grounding LLMs for task planning can effectively handle complex tasks, outperforming traditional planning methods and highlighting the potential for neuro-symbolic approaches in robotics.

Further investigation into construction applications is documented in the following sub-section, where the LLM agents are leveraged to aid construction tasks.

\subsection{Large Language Models in construction applications}

He, C. et al. \cite{c12} developed an LLM agent to classify and track constraints discussed in onsite construction planning meetings, improving the identification and management of issues like materials, labor, and equipment. By analyzing 94 meeting transcripts, the study reveals distinct discussion patterns between managers and foremen and demonstrates an 8.8 percent improvement in constraint classification accuracy using a GPT-based model and BERT.

Smetana, M. et al \cite{c13} utilized OpenAI's GPT-3.5 model and advanced natural language processing techniques to analyze highway construction safety incidents from OSHA’s Severe Injury Reports, aiming to improve accident cause identification. Their study analysis reveals major accident types, such as heat-related and struck-by injuries, and demonstrates the potential of AI in enhancing safety protocols and accident prevention strategies in the industry.

Regarding construction safety learning, Zaidi, S.F.A. et al. \cite{c14} introduced iSafe Chatbot, a novel framework for construction safety learning that used LLM agent to dynamically retrieve and explain relevant OSHA rules, while also recommending training videos based on user queries. By addressing limitations in traditional safety learning methods, iSafe Chatbot enhances engagement, comprehension, and adherence to safety protocols in the construction industry.

Prieto, S.A. et al. \cite{c8} developed a multi-LLM agent framework that uses GPT-4 LLM to optimize task allocation by integrating construction robots and human users, balancing time efficiency and resource usage. GPT-4 LLM agent was used to generate a series of commands after being provided with the complete API and the desired task objective. Their framework was tested in a simple scenario and demonstrated the potential of LLM agents to improve task scheduling and adapt to unforeseen site conditions, marking a significant advancement in AI-driven construction operations.


Prior work explores LLMs for task allocation and construction, but none evaluate a hybrid pipeline that combines GPT-4 with Gemma-3, LLaMA-4, and Mistral-7B or benchmark accuracy with standardized metrics. We propose a generator–supervisor hybrid and validate schedules using both plan-form and text-form metrics.


\section{Problem Formulation}
\begin{figure*}[!t]
    \centering
    \includegraphics[width=0.9\linewidth]{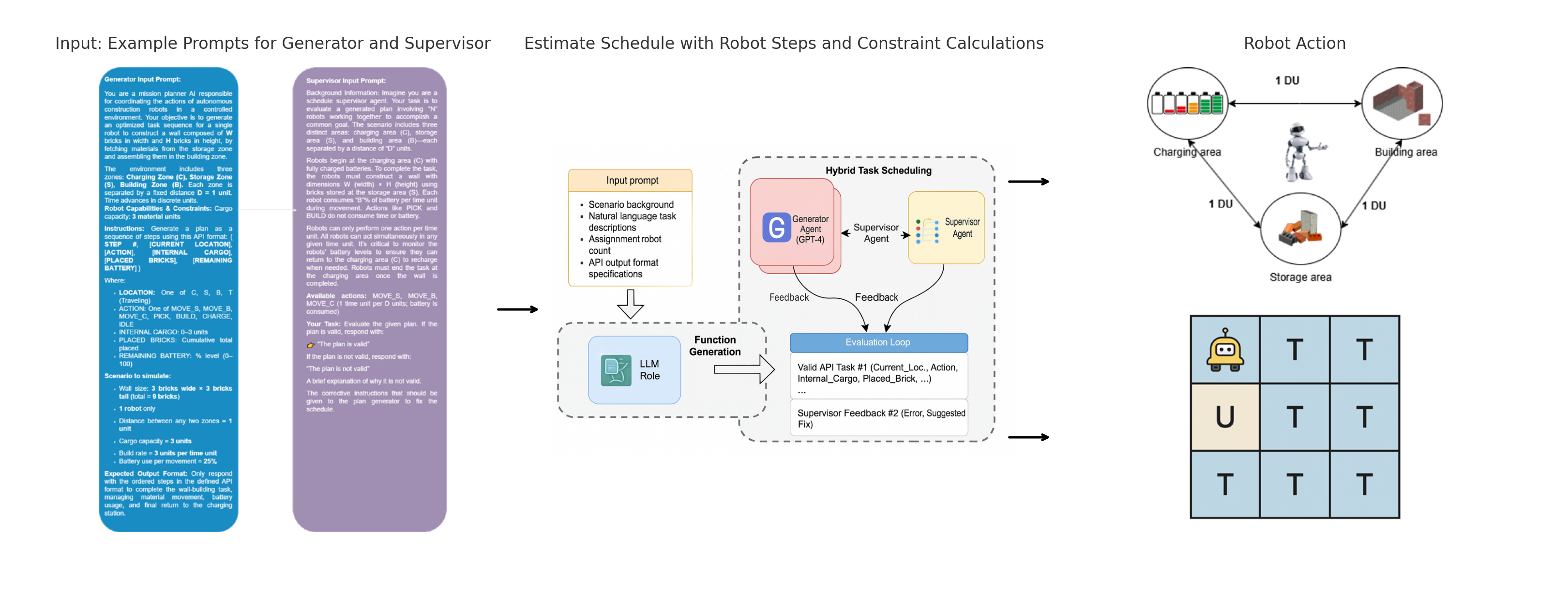}
    \caption{Hybrid LLM-Agent Framework for Multi-Robot Task Scheduling. The system uses GPT-4 as a Generator Agent and a second LLM (Gemma, LLaMA 4, or Mistral 7B) as a Supervisor Agent. Input prompts are iteratively refined via feedback loops to ensure logical consistency, producing API-level commands for autonomous robot execution.}
    \label{fig:hybrid_framework}
\end{figure*}
\textbf{Environment and Entities.}
Let the workspace be a planar construction site $\mathcal{W}\subset\mathbb{R}^2$ with a finite set of labeled locations $\mathcal{L}$ (e.g., \texttt{Stockpile\_A}, \texttt{Wall\_S1}, \texttt{Inspection\_Zone}). The site contains objects and resources $\mathcal{O}$ (e.g., bricks, markers, inspection targets) and static constraints (no-go zones, aisles) encoded as predicates over $(\mathcal{W},\mathcal{L})$.

\textbf{Robots and Capabilities.}
We consider $N$ heterogeneous autonomous robots
$\mathcal{R}=\{r_1,\ldots,r_N\}$.
Each robot $r_i$ is characterized by:
(i) a skill set $\Delta_i\subseteq\Delta$ over primitive actions $\Delta$ (e.g., \textsc{Navigate}, \textsc{Pick}, \textsc{Place}, \textsc{MarkLayout}, \textsc{Inspect}),
(ii) payload capacity $C_i$, battery state $B_i\in[0,B_i^{\max}]$,
(iii) current location $\ell_i\in\mathcal{L}$ and internal cargo state $c_i$.
A coalition is a nonempty subset $A\subseteq\mathcal{R}$ with aggregate skills $\Delta_A=\bigcup_{r_i\in A}\Delta_i$.

\textbf{High-Level Instruction and Tasks.}
A user provides an English instruction $X$ (e.g., ``survey, lay out, transport bricks, assemble, inspect'').
From $X$ and the scenario context we derive a finite set of atomic tasks
$\mathcal{T}=\{\tau_1,\ldots,\tau_M\}$, each with attributes
\[
\begin{aligned}
\big(&\mathrm{type}(\tau),\ \sigma(\tau)\subseteq\Delta,\ \ell(\tau)\in\mathcal{L},\\
&d(\tau)\in\mathbb{N}_{0},\ \mathrm{dur}(\tau)\in\mathbb{R}_{>0}\big).
\end{aligned}
\]
where $\sigma(\tau)$ are required skills, $\ell(\tau)$ is the target location, $d(\tau)$ is resource demand (e.g., brick count), and $\text{dur}(\tau)$ is a nominal duration.
Precedence constraints are given by a DAG $\mathcal{G}=(\mathcal{T},\prec)$; $\tau_i\prec\tau_j$ implies $\tau_i$ must complete before $\tau_j$ begins. Spatial, resource, and safety constraints are captured by a feasibility predicate $\Phi$ over plans (defined below).

\textbf{Schedule and API-Level Commands.}
Execution proceeds in discrete steps $k=1,\ldots,K$. A plan is a sequence
$\pi=\langle s_1,\ldots,s_K\rangle$ where each step is

\[
s_k =
\biggl\langle
\begin{array}{l}
\texttt{STEP}=k, \ \texttt{CURRENT\_LOCATION}=\ell_k,\\
\texttt{ACTION}=a_k, \ \texttt{INTERNAL\_CARGO}=c_k,\\
\texttt{PLACED\_BRICKS}=p_k, \\ \texttt{REMAINING\_BATTERY}=b_k
\end{array}
\biggr\rangle.
\]

with $a_k\in\Delta$ and state components consistent with robot/team dynamics. Let $\Pi$ denote the set of all syntactically valid command sequences under the API schema.

\textbf{Assignment and Timing.}
Let $\alpha:\mathcal{T}\rightarrow 2^{\mathcal{R}}\setminus\{\varnothing\}$ map each task to either a single robot or a coalition; $\alpha(\tau)=A$ is admissible iff $\sigma(\tau)\subseteq \Delta_A$ and all coalition constraints (e.g., payload sharing) hold. Let $\theta:\mathcal{T}\rightarrow \mathbb{R}_{\ge0}$ denote start times. A schedule $(\alpha,\theta)$ is \emph{feasible} if:
\begin{enumerate}
  \item \textbf{Precedence:} $\tau_i\prec\tau_j \Rightarrow \theta(\tau_j)\ge \theta(\tau_i)+\text{dur}(\tau_i)$.
  \item \textbf{Capability:} $\sigma(\tau)\subseteq\Delta_{\alpha(\tau)}$ for all $\tau$.
  \item \textbf{Resource/Capacity:} loads per robot respect $C_i$; brick placement counters update $p_k$ consistently with $d(\tau)$.
  \item \textbf{Battery and Kinematics:} for each $r_i$, cumulative motion/actuation implied by $\pi$ yields $b_k\in[0,B_i^{\max}]$ and realizable motions in $\mathcal{W}$.
  \item \textbf{Safety and Site Rules:} $\Phi(\pi)=\text{true}$ (no collisions; no-go constraints respected).
\end{enumerate}

\textbf{Hybrid LLM–Agent Problem.}
Let $C$ denote the structured \emph{context} (background scenario, task text, robot roster $\mathcal{R}$, API schema, samples). We model two operators:
\[
\begin{aligned}
\mathcal{F}_{\text{gen}} &: C \to \tilde{\pi},\\
\mathcal{F}_{\text{sup}} &: (\tilde{\pi}, \mathcal{G}, \mathcal{R}, \Phi) \to \hat{\pi}\in \Pi\cup\{\bot\}.
\end{aligned}
\]
where $\tilde{\pi}$ is a preliminary plan emitted by the \emph{Generator Agent} (GPT-4) and $\hat{\pi}$ is either a validated/corrected plan or the infeasibility token $\bot$ returned by the \emph{Supervisor Agent} (Gemma~3, LLaMA~4, or Mistral~7B). The supervisor enforces feasibility by projecting onto the valid plan set
\[
\mathcal{P}\triangleq
\Bigl\{\ \pi\in\Pi \ \Bigm|\ 
\begin{array}{l}
\exists(\alpha,\theta)\ \text{s.t.}\ \pi\ \text{realizes }(\alpha,\theta),\\
\text{and constraints 1--5 hold}
\end{array}
\Bigr\}.
\]
Let $\mathcal{V}(\pi)$ be a vector of violation indicators (precedence, capability, capacity, battery, safety), and let $\Psi(\pi)=\|\mathcal{V}(\pi)\|_0$ be a plan invalidity score. The supervisor seeks a minimal-edit correction
\[
\hat{\pi} \;=\; \arg\min_{\pi\in\mathcal{P}} \ \mathsf{dist}(\pi,\tilde{\pi}) \quad \text{subject to}\quad \Psi(\pi)=0,
\]
where $\mathsf{dist}(\cdot,\cdot)$ measures edits such as reordering, action substitution, coalition reassignment, or parameter fixes.

\textbf{Objectives.}
Primary objective is \emph{feasibility} (find $\hat{\pi}\in\mathcal{P}$). Secondary objectives guide tie-breaking:
(i) minimize makespan $\max_\tau \big(\theta(\tau)+\text{dur}(\tau)\big)$,
(ii) maximize robot utilization (parallelism consistent with $\mathcal{G}$),
(iii) minimize battery usage and travel,
(iv) minimize edit distance $\mathsf{dist}(\hat{\pi},\tilde{\pi})$ (i.e., least intrusive supervision).

\textbf{Baseline for Comparison.}
For benchmarking, the FCFS rule-based scheduler is the mapping
\[
\mathcal{F}_{\text{FCFS}}:(X,\mathcal{G},\mathcal{R})\ \mapsto\ \pi_{\text{seq}},
\]
which enforces $\prec$ and assigns the earliest available robot with $\sigma(\tau)\subseteq\Delta_i$, executing tasks sequentially (no coalition formation, no corrective feedback).

\textbf{Problem Statement.}
Given $(X,\mathcal{W},\mathcal{L},\mathcal{O},\mathcal{R},\mathcal{G},\Phi)$ and the API schema, compute a validated command sequence $\hat{\pi}\in\mathcal{P}$ that realizes an assignment $(\alpha,\theta)$ and optimizes the secondary criteria. We implement this as
\[
\hat{\pi}=\mathcal{F}_{\text{sup}}\bigl(\mathcal{F}_{\text{gen}}(C),\mathcal{G},\mathcal{R},\Phi\bigr).
\]

\section{METHODOLOGY}

 \subsection{System Overview}
Our pipeline converts an English instruction for a construction scene into an executable, API-level schedule via a two-agent loop: a \emph{Generator} (GPT-4) that proposes a plan and a \emph{Supervisor} (Gemma~3, LLaMA~4, or Mistral~7B) that validates and repairs it. The process is organized into staged prompts with code-like structure to enhance fidelity and executability (Figure~\ref{fig:hybrid_framework}). As in recent LLM planning systems, we use structured prompts and few-shot programmatic exemplars because they improve the model’s grasp of syntax and execution constraints, making the outputs easier to verify and run.

\subsection{Stage 0: Scenario Grounding and Canonicalization}
Given a user instruction and the site context, we assemble a canonical context block:
(i) background (site map, no-go zones, stockpiles), 
(ii) natural-language task list, 
(iii) robot roster with skills, capacities, and starting locations,
(iv) API schema and the target command format 
\texttt{\{STEP,\ CURRENT\_LOCATION,\ ACTION,\ INTERNAL\_CARGO,\ PLACED\_BRICKS,\ REMAINING\_BATTERY\}}.
Entity names (locations, robot IDs, materials) are normalized to a controlled vocabulary to reduce ambiguity during generation and checking.

\subsection{Stage 1: Generator Prompting (Plan Synthesis)}
We prompt the Generator with a compact, code-like template that embeds: (a) the canonical context, (b) few-shot examples of valid schedules, and (c) explicit schema and invariants (e.g., “respect precedence; do not duplicate actions; keep battery non-negative”). The Generator returns a preliminary schedule $\tilde{\pi}$ as a stepwise sequence of API commands. Using code-structured few-shot prompts (with comments and dictionary-like structures) helps the model map natural language tasks into executable steps while controlling token footprint.

\subsection{Stage 2: Supervisor Validation (Constraint Checking)}
The Supervisor receives $\tilde{\pi}$ and independently recomputes constraint satisfaction against the grounded scenario. We formalize five classes of checks:
\begin{enumerate}
    \item \textbf{Precedence:} all DAG constraints are respected; no task begins before its prerequisites finish.
    \item \textbf{Capability/Coalition:} each step’s \texttt{ACTION} is assigned to a robot (or team) with the required skills; if a single robot is insufficient, the Supervisor may authorize a minimal coalition.\footnote{Coalitions are allowed only when an action explicitly supports multi-robot execution (e.g., long-span carries).}
    \item \textbf{Capacity/Resources:} payload limits, brick counts, and placement tallies are consistent.
    \item \textbf{Battery/Kinematics:} travel and actions decrement battery; locations change only via navigation actions; battery never negative.
    \item \textbf{Schema/Uniqueness:} commands conform to the API; no duplicated or contradictory steps.
\end{enumerate}
Each violation emits a typed error with an actionable fix suggestion (e.g., “swap steps 5 and 6,” “split carry into two-robot coalition,” “insert navigate-to-stockpile before pick”).

\subsection{Stage 3: Supervised Repair Loop (Minimal-Edit Projection)}
If $\tilde{\pi}$ fails validation, the Supervisor proposes a corrected plan $\pi'$ that minimally edits the original with respect to (i) action substitutions, (ii) step reordering, (iii) coalition reassignment, and (iv) parameter fixes (locations, counts). We iterate a bounded number of times or until zero violations remain. Intuitively, the Supervisor acts as a projection operator that maps a syntactically valid but inconsistent plan to the nearest feasible one under the scenario constraints.


\begin{algorithm}[t]
\caption{Hybrid LLM Supervision for Feasible Scheduling}
\label{alg:hybrid}
\begin{adjustbox}{scale=0.8,center}
\begin{minipage}{\linewidth}
\begin{algorithmic}[1]
\Require Context $C$, Generator $\mathcal{G}$, Supervisor $\mathcal{S}$, max iters $T$
\State $\tilde{\pi} \gets \mathcal{G}(C)$
\For{$t \gets 1$ \textbf{to} $T$}
  \State $(ok,\, \mathcal{E}) \gets \textproc{validate}(\mathcal{S}, \tilde{\pi}, C)$
  \If{$ok$}
    \State \Return $\tilde{\pi}$
  \EndIf
  \State $\tilde{\pi} \gets \textproc{repair}(\mathcal{S}, \tilde{\pi}, \mathcal{E}, C)$
\EndFor
\State \Return \textsc{Infeasible}
\end{algorithmic}
\end{minipage}
\end{adjustbox}
\end{algorithm}

\subsection{Stage 4: Finalization and Execution Interface}
On success, the Supervisor emits a \emph{validated} schedule $\hat{\pi}$ plus a short justification log that references specific constraints it enforced. The schedule is already in API-ready form; an adapter translates each step into robot middleware calls (e.g., navigation, pick/place, layout marking, inspection). Where multiple sub-tasks are independent, the Supervisor may schedule them in parallel across robots, provided safety zones do not conflict. The staged, plan–coalition–allocation organization follows the general pattern of multi-stage LLM planners but is tailored to construction tasks and our API schema.

\subsection{Prompt Engineering and Guardrails}
Prompts are \emph{structured} rather than free-form prose: commented blocks, key–value lists for robots/objects, and explicit “do/don’t” rules. This format consistently improves the LLM’s adherence to syntax and reduces post-hoc repairs. We keep temperatures low for determinism, add stop tokens at schema boundaries, and include compact counterexamples (e.g., “duplicated action” cases) to prime the Supervisor’s error detection. These choices reflect evidence that code-like, line-commented exemplars enhance executability and comprehension.

\subsection{Coalitions and Parallelism}
When a sub-task requires skills or payload beyond a single robot, the Supervisor forms a minimal coalition and splits the action into synchronized primitives (e.g., \textsc{Navigate} $\rightarrow$ \textsc{CoCarry} $\rightarrow$ \textsc{Place}). Independent sub-tasks with equal temporal precedence may be assigned to distinct robots and executed concurrently, subject to safety buffers. This mirrors the general idea that some sub-tasks can be performed in parallel if precedence allows.


\subsection{Complexity and Cost Considerations}
Validation is linear in the number of steps (schema checks and simple counters) plus graph lookups for precedence. The dominant cost is LLM inference: one Generator call and up to $T$ Supervisor calls per scenario. The structured prompts and minimal-edit repairs are designed to keep $T$ small in practice.

\section{Experiments}

To directly align the empirical evidence with our framework’s multi-robot, coalition-capable formulation, we begin the evaluation with a cooperative, multi-agent scheduling scenario that explicitly exercises (i) coalition requirements, (ii) heterogeneous skills, (iii) precedence constraints, and (iv) safe parallel execution under shared resource constraints. We then report a controlled single-robot benchmark to isolate the effect of the proposed LLM--supervisor loop under minimal confounds. Finally, we include additional multi-agent case studies across multiple layout variants to illustrate how the same constraint-aware scheduling mechanism generalizes to different site geometries and interaction patterns.

Our framework uses a two-tier LLM pipeline: a Generator (GPT-4) produces a draft schedule from a structured prompt (scenario context, natural-language tasks, robot IDs, API spec), and a Supervisor (Gemma-3, LLaMA-4, or Mistral-7B) validates typed constraints (ordering, redundancy, feasibility) and applies minimal edits in a short repair loop until the plan is feasible. The validated schedule is then compiled to low-level robot commands. Each step follows the corresponding API schema.

We evaluate on two simulated construction tasks with explicit resource/safety constraints, comparing (i) Generator-only (no validation), (ii) the Hybrid generator+supervisor approach, and (iii) an FCFS rule-based baseline (precedence only; no semantic repair/coalitions). We report Feasibility Rate (FR), text similarity (BLEU, ROUGE-1/2/L, METEOR) between the Generator draft and the corrected plan, and an edit profile (insertions, reorders, substitutions). For plan correctness we also track Feasible Plan Rate (FPR), Edit Distance (\#Edits), battery-safety violations, makespan (TU), and supervisor iterations ($T_{\text{rep}}$). All methods use identical context (maps, costs, roster) and the same few-shot API schema; the hybrid loop is capped at $T=3$ repairs with low temperature, matched prompt budgets, and repair hints limited to typed violations.


\subsection{Experiment I: Battery-Constrained Wall Assembly}
\begin{figure}[t]
  \centering
  \includegraphics[width=0.9\linewidth]{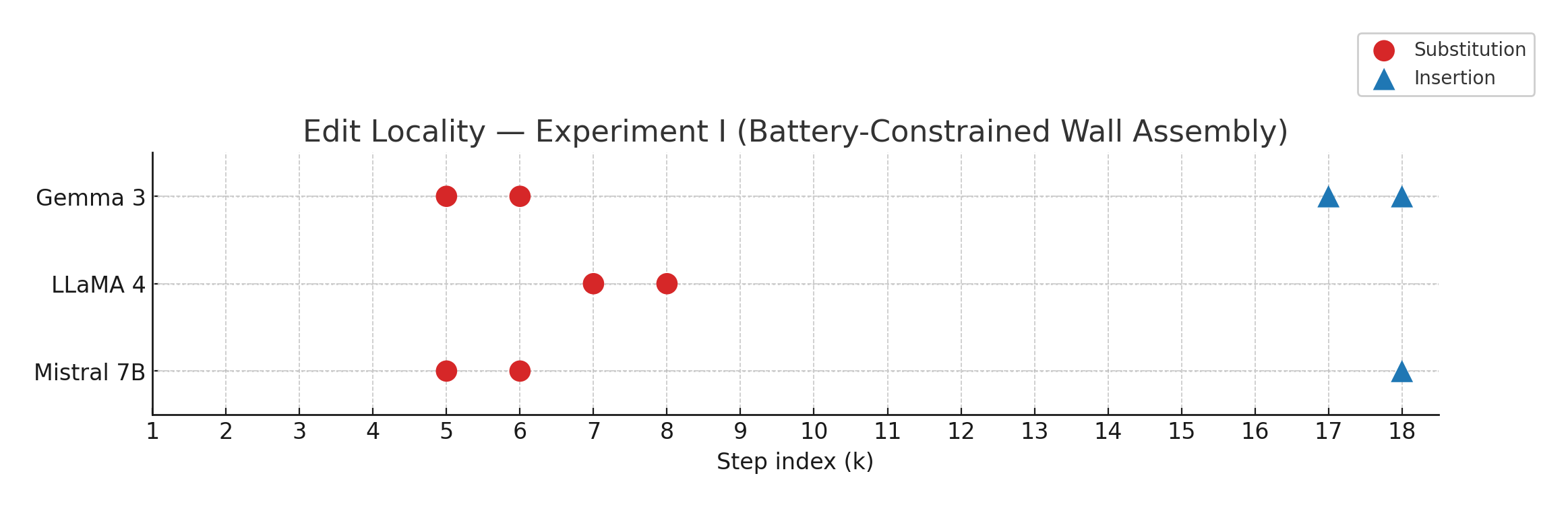}
  \caption{Edit locality for Experiment I (9-brick wall). Red = substituted steps; Blue = inserted steps. Edits cluster in a 2-step window around the battery violation (Gemma: S5–S6 + safe tail; LLaMA: delayed charge at S7–S8; Mistral: early charge + extra terminal charge).}
  \label{fig:edit_overlay_exp1}
\end{figure}

A single mobile robot must build a 9-brick wall by shuttling materials from a storage area to a build area with a charging station available; costs are specified in generic units (distance DU, time TU, materials MU), with battery usage at 25\%/DU, \textsc{Pick}/\textsc{Build} at 1~TU per 3~MU, and full recharge at 1~TU. The Generator's initial plan typically violates battery constraints (e.g., moving with $0\%$ battery or without a feasible return-to-charge). Supervisors correct by inserting \textsc{Move\_C}/\textsc{Charge} before critical moves and by fixing post-build navigation order, yielding FR$=1.0$ for all three models. This reduces battery violations to zero and preserves the original task decomposition. Edit locality is concentrated around the violation; see Fig.~\ref{fig:edit_overlay_exp1}. Gemma~3 typically performs a single insertion with minimal reordering; LLaMA~4 produces a near-identical fix; Mistral~7B sometimes adds a conservative extra charge, increasing edit distance without affecting FR.

\begin{table}[h]
\centering
\caption{Experiment I — Similarity between Generator and Corrected Plans}
\label{tab:exp1}
\begin{tabular}{lcccc}
\toprule
\textbf{Supervisor} & \textbf{BLEU} & \textbf{ROUGE-1} & \textbf{ROUGE-L} & \textbf{METEOR} \\
\midrule
Gemma 3   & 0.9407 & 0.9444 & 0.9444 & 0.9655 \\
LLaMA 4   & 0.8750 & 0.9230 & 0.9230 & 0.9140 \\
Mistral 7B & 0.8235 & 0.8824 & 0.8235 & 0.8529 \\
\bottomrule
\end{tabular}
\end{table}

\subsection{Experiment II: Scan Coverage and Path Feasibility}
\begin{figure}[t]
  \centering
  \includegraphics[width=0.9\linewidth]{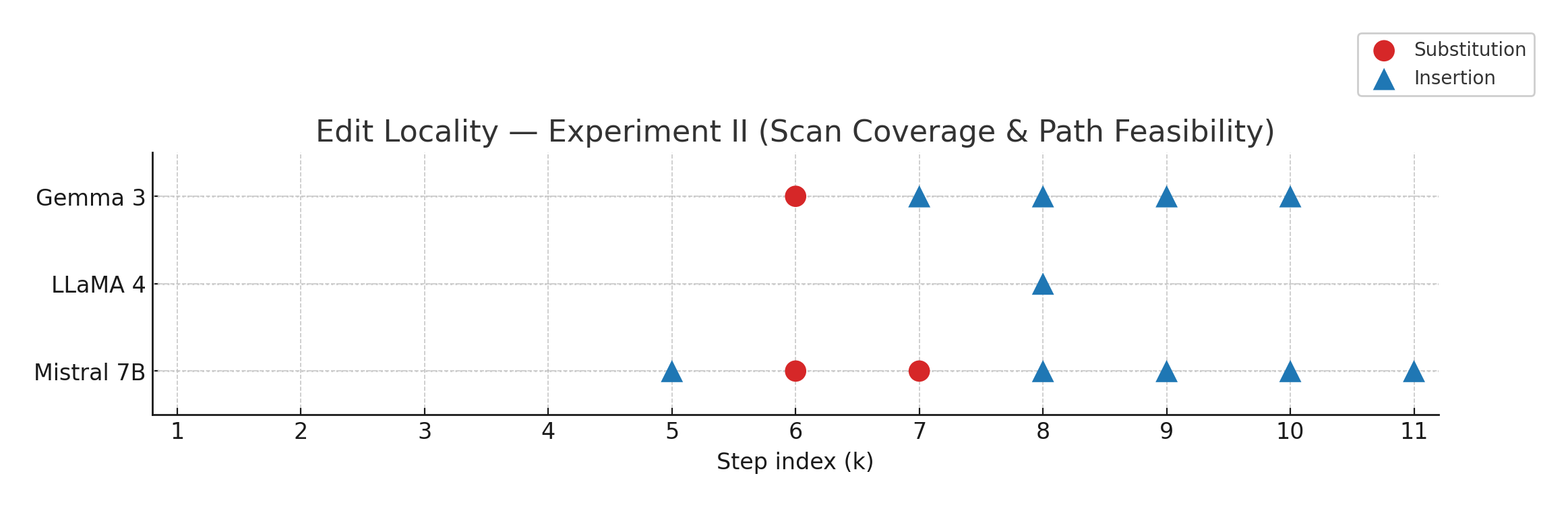}
  \caption{Edit locality for Experiment II (scan coverage). LLaMA fixes coverage with a single \textsc{SCAN} insertion (S8); Gemma performs a short move$\to$scan detour (S6–S9); Mistral adds redundant scans and moves (S5, S8–S11) to build extra sensing margin.}
  \label{fig:edit_overlay_exp2}
\end{figure}

A robot must (i) discover/scan each traversable grid cell exactly once and (ii) return a valid path from $(0,0)$ to $(2,2)$ while respecting a blocked cell and battery usage at 15\%/DU; scan cost is 1~TU/SU and recharge is 1~TU.The Generator often leaves a corner cell unscanned (e.g., $(2,0)$). All Supervisors reach FR$=1.0$ by detecting the coverage gap and inserting a strategically placed \textsc{Scan} (e.g., at $(1,1)$ or $(2,2)$) with the minimal detour to ensure full discovery, then regenerate the final path. Step-level repairs are local; see Fig.~\ref{fig:edit_overlay_exp2}. LLaMA~4 yields the most \emph{surgical} correction (single additional \texttt{SCAN} whose footprint completes the map), whereas Mistral~7B often appends extra steps (lower BLEU) while preserving most of the original content (high ROUGE/METEOR). Gemma~3 completes coverage but modifies a longer prefix, reducing n-gram precision. Hybrid methods consistently achieve full coverage and a valid path (FPR~=~100\%), with $\#\text{Edits}$ dominated by a single \textsc{Scan} insertion and a small reroute. Generator-only fails coverage checks in a sizable fraction of trials; FCFS cannot reason about discovery semantics, yielding lower FPR.

\begin{table}[t]
\centering
\caption{Experiment II — Similarity between Generator and Corrected Plans}
\label{tab:exp2}
\renewcommand{\arraystretch}{1.15}
\begingroup
\setlength{\tabcolsep}{3pt}           
\fontsize{9pt}{10.5pt}\selectfont     
\begin{tabular}{lccccc}
\toprule
\textbf{Supervisor} & \textbf{BLEU} & \textbf{R-1} & \textbf{R-2} & \textbf{R-L} & \textbf{METEOR} \\
\midrule
Gemma 3    & 0.447  & 0.625  & --    & 0.625  & 0.672 \\
LLaMA 4    & 0.742  & 0.933  & 0.923 & 0.933  & 0.984 \\
Mistral 7B & 0.339  & 0.7778 & 0.625 & 0.7778 & 0.934 \\
\bottomrule
\end{tabular}
\endgroup
\end{table}

\subsection{Baselines}
We benchmark three planners under identical setup (API schema, maps/costs/robot roster, few-shot prompts, and token budget): 
\textbf{(i) Generator-only (GPT-4)}—one low-temperature pass ($\le 0.2$; fixed seed; schema stop-tokens); the draft is used as-is with no validation or repair. 
\textbf{(ii) Hybrid}—a \textbf{Generator} (GPT-4) proposes a schedule that a \textbf{Supervisor} (Gemma~3, LLaMA~4, or Mistral~7B) checks using only the same context and the draft; it applies typed fixes (schema, precedence, capability/capacity, battery, coverage) via minimal insert/reorder/substitute edits for up to \(T=3\) validate$\rightarrow$repair loops. 
\textbf{(iii) FCFS rule-based}—deterministic topological scheduling that assigns each task to the earliest feasible robot; no coalitions, no battery/coverage reasoning, and no post-hoc edits. 
All scripts assume a unit-cost travel/action model and are emitted in the same API format.

\section{Analysis}

We analyze how supervision converts high-likelihood but infeasible drafts into executable schedules by inspecting \emph{where} and \emph{how} each supervisor modifies the Generator’s step trace. Our discussion triangulates text-form metrics (BLEU/ROUGE/METEOR) with plan-form evidence (battery/coverage violations, edit types, and makespan).

\subsection{Text-Form vs. Plan-Form Signals}
Across both tasks, feasibility (FR) hinges on a \emph{small, typed} change: one strategically placed \textsc{CHARGE} for energy feasibility or one \textsc{SCAN} at a cell whose footprint closes the coverage gap. Text similarity then reflects how minimally the supervisor deviates from the draft: higher BLEU/ROUGE/METEOR indicates fewer edits and tighter alignment; lower BLEU with high ROUGE/METEOR indicates recall-heavy fixes that preserve content but insert extra steps.

\subsection{Experiment I: Battery-Constrained Wall Assembly}
\begin{figure}[t]
  \centering
  \includegraphics[width=0.9\linewidth]{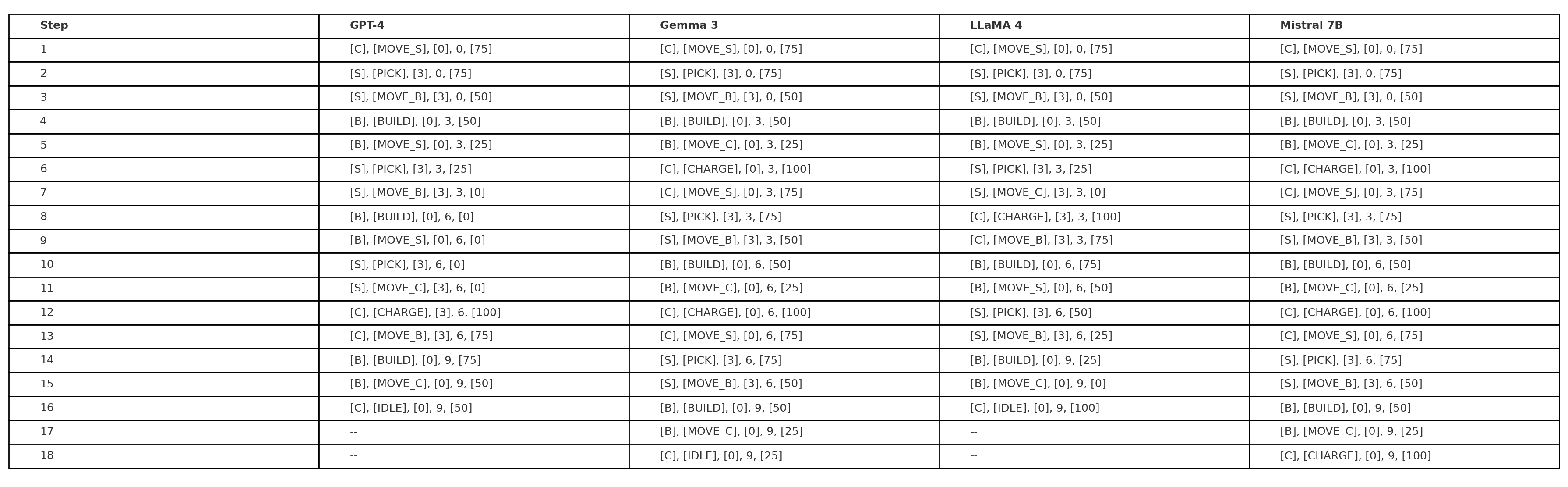}
  \caption{Experiment I: 9-Brick Wall Assembly}
  \label{fig:exp1steps}
\end{figure}

\paragraph{Failure mode in the draft.} In Figure~\ref{fig:exp1steps}, the Generator runs a classic energy shortfall: after \texttt{STEP 4} (\texttt{[B], BUILD, [0], 3, [50]}), it goes \texttt{[B]$\to$[S]} (\texttt{STEP 5}) and \texttt{PICK} at \texttt{STEP 6} with only 25\% battery remaining, then attempts \texttt{[S]$\to$[B]} at \texttt{STEP 7} and arrives at \texttt{[B]} with \texttt{0\%} battery. The same pattern reoccurs around \texttt{STEP 9}–\texttt{STEP 11}. Precedence holds, but the plan violates energy feasibility (insufficient margin to reach, build, or return to charge).

\paragraph{Gemma~3: minimal edit with early charge.}
Gemma rewires the first post-build transition to go \texttt{[B]$\to$[C]} (substitution at \texttt{STEP 5}: \texttt{MOVE\_S}$\Rightarrow$\texttt{MOVE\_C}) and \emph{replaces} the risky \texttt{PICK} with \texttt{CHARGE} (substitution at \texttt{STEP 6}: \texttt{PICK}$\Rightarrow$\texttt{CHARGE}, battery $\to 100$). It then resumes the material shuttle by going \texttt{[C]$\to$[S]} at \texttt{STEP 7} (\texttt{[75]\%} remaining), preserving the original decomposition of trips. At the tail, Gemma appends two safe-guarding steps (\texttt{STEP 17: [B]$\to$[C]}, \texttt{STEP 18: IDLE}), yielding a slightly longer but conservative wind-down. \textbf{Ablation (battery check only):} these two early substitutions alone remove all battery violations and are sufficient for feasibility; the tail additions only add a small makespan overhead (idle/return-to-charge policy).

\emph{Edit profile}: 2 substitutions (S5, S6) + 2 insertions (S17, S18); \#reorders $=0$.  
\emph{Effect}: FR$\uparrow$ to 1.0; BLEU/ROUGE remain very high (Table~\ref{tab:exp1}); makespan $\approx$ draft $+$(1–2)~TU.

\paragraph{LLaMA~4: surgical charge with delayed timing.}
LLaMA keeps \texttt{S5: [B]$\to$[S]} intact but routes \texttt{S7: [S]$\to$[C]} (instead of \texttt{[S]$\to$[B]}), arriving with \texttt{0\%} battery (non-negative edge case), then \texttt{S8: CHARGE} to \texttt{100\%}. Because it defers charging, the edit shows up as a \emph{local reorder} (swap the “pick” and “charge” windows) rather than an early substitution at S5–S6. Downstream, LLaMA preserves the remaining structure, only altering a few navigation steps (e.g., \texttt{S11: [B]$\to$[S]}) to keep the shuttle cadence. \textbf{Ablation (battery check only):} a single additional \texttt{CHARGE} (at S8) suffices to eliminate the underflow; no extra terminal steps are needed.

\emph{Edit profile}: 1 reorder window (S7–S8), $\leq$2 substitutions total; \#insertions $=0$.  
\emph{Effect}: FR$=1.0$; near-draft BLEU/ROUGE (Table~\ref{tab:exp1}); makespan $\approx$ draft $+1$~TU.

\paragraph{Mistral~7B: conservative margin with extra charge.}
Like Gemma, Mistral switches \texttt{S5: [B]$\to$[C]} and \texttt{S6: CHARGE}, then mirrors Gemma’s mid-plan flow but \emph{also} adds a late \texttt{CHARGE} (\texttt{S18}) after completion. This “belt-and-suspenders” style creates more edits and slightly longer scripts without affecting feasibility.

\emph{Edit profile}: 2 substitutions (S5, S6) + 1 insertion (S18); \#reorders $=0$.  
\emph{Effect}: FR$=1.0$; lowest BLEU among supervisors due to extra step, yet ROUGE stays high (Table~\ref{tab:exp1}); makespan $\approx$ draft $+1$–$2$~TU.

Battery safety is a single-point failure corrected by a single charge window. \emph{Minimal-edit} supervisors (Gemma, LLaMA) stay closest to the draft (higher BLEU/ROUGE), whereas \emph{conservative} supervision (Mistral) inserts an extra safety cycle (lower BLEU, similar ROUGE). All three achieve FR$=1.0$ with $\leq2$ local changes to the early shuttle loop. 

\subsection{Experiment II: Scan Coverage and Path Feasibility}
\begin{figure}[t]
  \centering
  \includegraphics[width=0.9\linewidth]{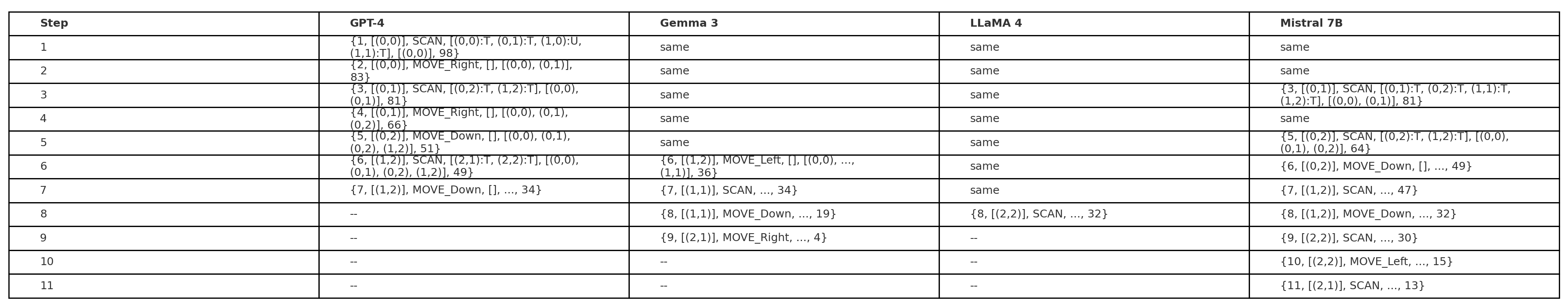}
  \caption{Experiment II: Scan Coverage \& Path Feasibility}
  \label{fig:exp2steps}
\end{figure}

\paragraph{Failure mode in the draft.} In Figure~\ref{fig:exp2steps}, the Generator leaves $(2,0)$ undiscovered. The path from $(0,0)$ to $(2,2)$ is plausible, but coverage is incomplete (a semantic, not syntactic, violation).

\paragraph{Gemma~3: re-centering the sensor footprint.}
Gemma converts the \texttt{S6} \texttt{SCAN} into a repositioning step (\texttt{MOVE\_Left}) to reach \texttt{(1,1)}, then performs a \texttt{SCAN} at \texttt{S7}. It follows with \texttt{MOVE\_Down} and \texttt{MOVE\_Right} (S8–S9) to keep the traversal compact while ensuring the \emph{sensing footprint} reveals the missing cell. This is a \emph{move→scan} micro-macro swap intended to scan from a more informative vantage point.

\emph{Edit profile}: $\approx$1 substitution (S6), 1 insertion (S7), 2 short moves (S8–S9); local reorder window length $\leq2$.  
\emph{Effect}: FR$=1.0$; moderate BLEU due to movement detour; ROUGE/METEOR stable (Table~\ref{tab:exp2}); makespan $+3$–$4$~TU (two moves + one scan).

\paragraph{LLaMA~4: one-shot scan at the goal corner.}
LLaMA leaves S1–S7 untouched and inserts a single \texttt{SCAN} at \texttt{S8} at \texttt{(2,2)}, whose adjacency footprint closes $(2,0)$. No extra repositioning is required; the remainder matches the draft trajectory. This is the most \emph{surgical} fix: a single, well-placed perception action.

\emph{Edit profile}: 1 insertion (S8); no reorders; no substitutions.  
\emph{Effect}: FR$=1.0$; top BLEU/ROUGE/METEOR (Table~\ref{tab:exp2}); makespan $+1$~TU.

\paragraph{Mistral~7B: redundant scans for robustness.}
Mistral adds an early \texttt{SCAN} at \texttt{S5} and later scans at \texttt{S9} and \texttt{S11}, plus extra moves (\texttt{S6}, \texttt{S10}). The policy is explicitly \emph{recall-seeking}: scan from multiple adjacent cells to ensure coverage even under partial observability or mis-estimated footprints. The result is a longer script with substantially lower BLEU but high ROUGE/METEOR (the corrected plan contains most of the draft’s $n$-grams plus additional, recall-oriented content).

\emph{Edit profile}: 3 insertions (S5, S9, S11), 2 move substitutions (S6, S10); local reorders around the new scans.  
\emph{Effect}: FR$=1.0$; BLEU lowest, ROUGE/METEOR high (Table~\ref{tab:exp2}); makespan $+4$–$6$~TU.

Coverage is a \emph{local sensing} problem: one well-placed \texttt{SCAN} suffices. LLaMA’s single insertion yields the best lexical alignment and minimal overhead; Gemma pays a small detour to scan from a more central cell; Mistral purchases robustness with redundant sensing (lower BLEU, higher makespan) while keeping FR$=1.0$.


\subsection{Guardrail-Level Ablation from Step Traces}
The step traces allow a concrete guardrail ablation:

\begin{itemize}
  \item \textbf{Schema/precedence only} (no semantics): leaves both drafts infeasible (battery underflow; uncovered cell). The traces show well-formed command strings but unchanged failure modes.
  \item \textbf{Battery safety only} (Exp.~I): a \emph{single} \texttt{CHARGE} window (Gemma S5–S6; LLaMA S8) eliminates all $b_k\!\le\!0$ events with 0–2 local reorders; makespan overhead $\leq 2$~TU.
  \item \textbf{Coverage completeness only} (Exp.~II): a \emph{single} \texttt{SCAN} (LLaMA S8) or a short move→scan detour (Gemma S6–S9) closes the map gap; makespan $+1$–$4$~TU depending on detour.
  \item \textbf{Battery + Coverage} (full supervisor): combines the above with optional terminal safety (e.g., \texttt{IDLE}/\texttt{CHARGE}); feasibility on both tasks with minimal edits and small overhead.
\end{itemize}

A consolidated edit-profile ablation across both tasks is reported in Table~\ref{tab:ablation_combined_edits}.

\begin{table*}[!t]
\centering
\caption{Combined Edit Profile Ablation — Experiment I (Battery-Constrained Wall Assembly) and Experiment II (Scan Coverage \& Path Feasibility)}
\label{tab:ablation_combined_edits}
\renewcommand{\arraystretch}{1.15}
\resizebox{\textwidth}{!}{
\begin{tabular}{l|cccccc|cccccc}
\toprule
& \multicolumn{6}{c|}{\textbf{Experiment I: Battery-Constrained Wall Assembly}} &
  \multicolumn{6}{c}{\textbf{Experiment II: Scan Coverage \& Path Feasibility}} \\
\textbf{Supervisor} &
\textbf{Minimal Edits} & \textbf{Edited Steps (minimal)} & \textbf{Additional/Policy Edits} &
$\boldsymbol{\Delta}$\textbf{Makespan (TU)} & \textbf{FR} & \textbf{Strategy} &
\textbf{Minimal Edits} & \textbf{Edited Steps (minimal)} & \textbf{Additional/Policy Edits} &
$\boldsymbol{\Delta}$\textbf{Makespan (TU)} & \textbf{FR} & \textbf{Strategy} \\
\midrule
\textbf{Gemma 3} &
Substitutions = 2,\; Insertions = 0,\; Reorders = 0 &
S5:\;MOVE\_S$\to$MOVE\_C;\; S6:\;PICK$\to$CHARGE &
Insertions = 2 (S17:\;MOVE\_C,\; S18:\;IDLE) &
$\sim$+1--2 & 1.0 & Minimal + conservative tail &
Substitutions = 1,\; Insertions = 1,\; Reorders = 0 &
S6:\;SCAN$\to$MOVE\_Left;\; S7:\;SCAN (+) &
Insertions = 2 (S8:\;MOVE\_Down,\; S9:\;MOVE\_Right) for short detour &
$\sim$+3--4 & 1.0 & Re-center sensor footprint (short detour) \\
\textbf{LLaMA 4} &
Substitutions = 2,\; Insertions = 0,\; Reorders = 0 &
S7:\;MOVE\_B$\to$MOVE\_C;\; S8:\;BUILD$\to$CHARGE &
None &
$\sim$+1 & 1.0 & Surgical minimal-edit &
Substitutions = 0,\; Insertions = 1,\; Reorders = 0 &
S8:\;SCAN (+) at (2,2) &
None &
$\sim$+1 & 1.0 & One-shot scan (surgical) \\
\textbf{Mistral 7B} &
Substitutions = 2,\; Insertions = 0,\; Reorders = 0 &
S5:\;MOVE\_S$\to$MOVE\_C;\; S6:\;PICK$\to$CHARGE &
Insertion = 1 (S18:\;CHARGE) &
$\sim$+1--2 & 1.0 & Conservative margin (extra charge) &
Substitutions = 2,\; Insertions = 1,\; Reorders = 0 &
S6:\;SCAN$\to$MOVE\_Down;\; S7:\;MOVE\_Down$\to$SCAN;\; S5:\;SCAN (+) &
Insertions = 3 (S9:\;SCAN,\; S10:\;MOVE\_Left,\; S11:\;SCAN) for redundancy &
$\sim$+4--6 & 1.0 & Redundant scans (recall-heavy) \\
\bottomrule
\end{tabular}}
\end{table*}

\subsection{Iteration Budget and Edit Locality}

All supervisors converge within a small loop ($T_{\text{rep}}\le 3$). Edits are \emph{local}: a two-step window around the violation is typical (S5–S6 in Experiment~I; S7–S8 in Experiment~II). This locality explains high ROUGE-L (longest common subsequence remains intact) even when BLEU drops due to extra tokens/steps.


Our results show that \emph{typed supervision} is an efficient and scalable mechanism for correcting LLM-generated plans in construction settings: enforcing a small set of typed checks—battery safety and coverage completeness—consistently drives feasibility to FR$=1.0$ with at most $\leq2$ localized edits across both tasks. Within this regime, supervisors exhibit distinct correction styles: Gemma and LLaMA act like projection operators onto the feasible set, introducing the minimum necessary change (typically a single \textsc{CHARGE} or \textsc{SCAN} insertion with a short local reorder), whereas Mistral adopts a conservative margin, adding redundant but harmless steps that slightly increase makespan while preserving feasibility. Crucially, because all supervision is applied as small, typed, and local modifications, the repaired schedules remain close to the Generator’s original intent, which improves traceability for human inspection and eases reliable handoff to robot middleware without additional engineering.

\section{CONCLUSION AND FUTURE WORK}

Supervision acts as a minimal-edit projection: it maps a well-formed but inconsistent draft to the nearest feasible schedule. When violations are local (e.g., one missing SCAN), LLaMA-4 produces the most surgical fix; more conservative supervisors add safety steps that lower BLEU but keep high ROUGE/METEOR and feasibility. The hybrid setup overcomes myopic next-token planning by separating generation and verification across LLM roles. Gemma-3 attains the best text-form scores (BLEU 0.9407, ROUGE-1 0.9444, METEOR 0.9655) and reliably validates/corrects plans, making GPT-4 + Gemma-3 a strong default pairing for efficient task scheduling. Future work will role-tune the agents and scale to heterogeneous multi-robot settings with specialized transport vs. assembly capabilities.

$$
$$

\addtolength{\textheight}{-12cm}   










\begin{thebibliography}{99}

\bibitem{c1} Zhao, S., Wang, Q., Fang, X., Liang, W., Cao, Y., Zhao, C., ... \& Wang, K. (2022). Application and development of autonomous robots in concrete construction: Challenges and opportunities. Drones, 6(12), 424.
\bibitem{c2} Team, G., Kamath, A., Ferret, J., Pathak, S., Vieillard, N., Merhej, R., ... \& Iqbal, S. (2025). Gemma 3 technical report. arXiv preprint arXiv:2503.19786.
\bibitem{c3} "The Llama 4 herd: The beginning of a new era of natively multimodal AI innovation | Meta." Accessed Apr. 08, 2025. [Online]. Available: https://ai.meta.com/blog/llama-4-multimodal-intelligence/
\bibitem{c4} Jiang, A. Q., Sablayrolles, A., Mensch, A., Bamford, C., Chaplot, D. S., Casas, D., ... \& Lavaud, L. Mistral 7b. arXiv [Preprint](2023). arXiv preprint arXiv:2310.06825.
\bibitem{c5} Achiam, J., Adler, S., Agarwal, S., Ahmad, L., Akkaya, I., Aleman, F. L., ... \& McGrew, B. (2023). Gpt-4 technical report. arXiv preprint arXiv:2303.08774.
\bibitem{c6} Kannan, S. S., Venkatesh, V. L., \& Min, B. C. (2024, October). Smart-llm: Smart multi-agent robot task planning using large language models. In 2024 IEEE/RSJ International Conference on Intelligent Robots and Systems (IROS) (pp. 12140-12147). IEEE.
\bibitem{c7} Jin, Y., Li, D., Shi, J., Hao, P., Sun, F., Zhang, J., \& Fang, B. (2024). Robotgpt: Robot manipulation learning from chatgpt. IEEE Robotics and Automation Letters, 9(3), 2543-2550.
\bibitem{c8} Prieto, S. A., \& Garcia de Soto, B. (2024, May). Large Language Models for Robot Task Allocation. In J. (mississippi S. U. Chen, Y. K. (georgia I. of T. Cho, I. (north D. S. U. Jeong, C. (new Y. U. Feng, B. (new Y. U. A. D. García de Soto, L. (baidu R. Zhang, … M. (hilti) Helmberger (Eds.), Proceedings of the 3rd Future of Construction Workshop at the International Conference on Robotics and Automation (ICRA 2024) (pp. 17–20). doi:10.22260/ICRA2024/0007
\bibitem{c9} Wang, J., \& Ke, L. (2024). Llm-seg: Bridging image segmentation and large language model reasoning. In Proceedings of the IEEE/CVF Conference on Computer Vision and Pattern Recognition (pp. 1765-1774).
\bibitem{c10} Wake, N., Kanehira, A., Sasabuchi, K., Takamatsu, J., \& Ikeuchi, K. (2024). Gpt-4v (ision) for robotics: Multimodal task planning from human demonstration. IEEE Robotics and Automation Letters.
\bibitem{c11} Chalvatzaki, G., Younes, A., Nandha, D., Le, A. T., Ribeiro, L. F., \& Gurevych, I. (2023). Learning to reason over scene graphs: a case study of finetuning GPT-2 into a robot language model for grounded task planning. Frontiers in Robotics and AI, 10, 1221739.
\bibitem{c12} He, C., Yu, B., Liu, M., Guo, L., Tian, L., \& Huang, J. (2024). Utilizing large language models to illustrate constraints for construction planning. Buildings, 14(8), 2511.
\bibitem{c13} Smetana, M., Salles de Salles, L., Sukharev, I., \& Khazanovich, L. (2024). Highway construction safety analysis using large language models. Applied Sciences, 14(4), 1352.
\bibitem{c14} ZAIDI, S. F. A., ABBAS, M. S., HUSSAIN, R., SABIR, A., Nasrullah, K. H. A. N., \& Jaehun, Y. A. N. G. (2024). iSafe Chatbot: Natural Language Processing and Large Language Model Driven Construction Safety Learning through OSHA Rules and Video Content Delivery. In International conference on construction engineering and project management (pp. 1238-1245). Korea Institute of Construction Engineering and Management. 
\bibitem{c15} Bernard, R., Raza, S., Das, S., \& Murugan, R. (2024). EQUATOR: A Deterministic Framework for Evaluating LLM Reasoning with Open-Ended Questions.\# v1. 0.0-beta. arXiv preprint arXiv:2501.00257.
\bibitem{c16} Xiong, G., Deng, Z., Wang, K., Cao, Y., Li, H., Yu, Y., ... \& Xie, Q. (2025). FLAG-Trader: Fusion LLM-Agent with Gradient-based Reinforcement Learning for Financial Trading. arXiv preprint arXiv:2502.11433.

\end{thebibliography}
\end{document}